\documentclass[12pt]{article}
\usepackage{times}  % DO NOT CHANGE THIS
\usepackage{helvet} % DO NOT CHANGE THIS
\usepackage{courier}  % DO NOT CHANGE THIS
\usepackage[hyphens]{url}  % DO NOT CHANGE THIS
\usepackage{graphicx} % DO NOT CHANGE THIS
\usepackage{graphicx}  % DO NOT CHANGE THIS
\usepackage{amsmath}
\usepackage{booktabs}
\usepackage{algorithm}
\usepackage{amssymb}
\usepackage{subfig}
\usepackage{array}
\usepackage{multirow}
\usepackage{placeins}
\usepackage{natbib}
\usepackage{hyperref}
\usepackage{algorithmic}
\renewcommand{\cite}{\citep}
\usepackage{bm}
\usepackage{enumitem}
\usepackage{extarrows}
\usepackage[american]{babel}

\usepackage[ruled,noresetcount,algo2e]{algorithm2e}
\newenvironment{sciabstract}{%
\begin{quote} \baselineskip14pt\small\hfil {\bf Abstract} \hfil\\[3pt]}
{\end{quote}\vspace{6pt}}

\DeclareMathOperator*{\argmax}{arg\,max}

\newcounter{lastnote}

\title{Derivative-Free Reinforcement Learning: A Review}

\author{
Hong Qian, Yang Yu$^{*}$\\
\normalsize{
National Key Laboratory for Novel Software Technology, Nanjing University, China}\\
\normalsize{
qianh@lamda.nju.edu.cn,\ yuy@nju.edu.cn}\\
\normalsize{$^*$Corresponding author.}
}

\date{}

\begin{document}

\baselineskip16pt

\maketitle 

\begin{sciabstract}
	 Reinforcement learning is about learning agent models that make the best sequential decisions in unknown environments. In an unknown environment, the agent needs to explore the environment while exploiting the collected information, which usually forms a sophisticated problem to solve. Derivative-free optimization, meanwhile, is capable of solving sophisticated problems. It commonly uses a sampling-and-updating framework to iteratively improve the solution, where exploration and exploitation are also needed to be well balanced. Therefore, derivative-free optimization deals with a similar core issue as reinforcement learning, and has been introduced in reinforcement learning approaches, under the names of \emph{learning classifier systems} and \emph{neuroevolution/evolutionary reinforcement learning}. Although such methods have been developed for decades, recently, derivative-free reinforcement learning exhibits attracting increasing attention. However, recent survey on this topic is still lacking. In this article, we summarize methods of derivative-free reinforcement learning to date, and organize the methods in aspects including parameter updating, model selection, exploration, and parallel/distributed methods. Moreover, we discuss some current limitations and possible future directions, hoping that this article could bring more attentions to this topic and serve as a catalyst for developing novel and efficient approaches.
\end{sciabstract}

\section{Introduction}
Reinforcement learning~\cite{sutton_reinforcement_1998,RL-SOTA-book} aims to enable agents to automatically learn the policy with the maximum long-term reward via interactions with environments. It has been listed as one of the four research directions of machine learning by Professor T. G. Dietterich~\cite{dietterich_machine_1997}. In recent years, with the fusion of deep learning and reinforcement learning, deep reinforcement learning has made remarkable progress and attracted more and more attention from both the academic and industrial community. To name a few, the deep Q-network (DQN)~\cite{mnih_human-level_2015} proposed by DeepMind reaches the human-level control in Atari games, AlphaGo~\cite{silver_mastering_2016} proposed also by DeepMind defeats the top human experts in the Go game, and AlphaZero~\cite{alphazero} defeats a world champion program in the games of chess, shogi and Go without the domain knowledge except the game rules. Due to the strong ability of reinforcement learning, it has been applied in automatic control~\cite{abbeel_application_2006}, automatic machine learning~\cite{iclr-ZophL17}, computer vision~\cite{iccv-HuangLR17}, natural language processing~\cite{aaai-YuZWY17}, scheduling~\cite{wang_application_2005}, finance~\cite{choi_reinforcement_2009}, commodity search~\cite{aaai-Shi0DCZ19}, and network communication~\cite{boyan_packet_1993}, etc. In the field of cognition and neuroscience, reinforcement learning also has important research value~\cite{frank_by_2004,samejima_representation_2005}.

However, as reinforcement learning is being applied to more realistic problems, the complexity of finding out an optimal or satisfactory policy is also increasing. The optimization problems encountered in reinforcement learning, especially deep reinforcement learning, are often quite sophisticated. For such optimization problems, standard gradient-based optimization methods may suffer from the difficulties of stationary point issues (e.g., a plethora of saddle points or spurious local optima), bad condition number, or flatness in the activations that could lead to the gradient vanishing problem~\cite{shalev-shwartz_failures_2017}. These difficulties cannot be neglected since they could result in the unsatisfactory performance of gradient-based optimization methods. Therefore, more effective policy learning methods that could make up for the shortcomings of gradient-based ones are quite appealing.

Derivative-free optimization~\cite{DFO-book,DFO-siamrev,DFO-review}, also termed as zeroth-order or black-box optimization, involves a kind of optimization algorithms that do not rely on the gradient information. Given a function $f$ defined over a continuous, discrete, or mixed search space $\mathcal{X}$, it only relies on the objective function value (or fitness value) $f(x)$ on the sampled solution $x$. Since the conditions of using derivative-free algorithms are relaxed, they are easy to use and suitable for dealing with the sophisticated optimization tasks, e.g., non-convex and non-differentiable objective functions. Moreover, derivative-free optimization commonly uses a sampling-and-updating framework to iteratively improve the quality of solutions. One of the key issues is to balance the exploration and exploitation in the search space. This key issue is in a quite similar situation of reinforcement learning. Therefore, the fusion of derivative-free optimization and reinforcement learning, termed as \emph{derivative-free reinforcement learning} in this paper, has many potentialities.

Derivative-free reinforcement learning has been developed for decades, under the names of \emph{learning classifier systems} (e.g.,~\cite{LCS-survey}) and \emph{neuroevolution/evolutionary reinforcement learning} (e.g.,~\cite{jair-MoriartySG99,Whiteson12}). Although it has some history, we have noticed that derivative-free reinforcement learning attracts increasing attentions recently. This article reviews some recent advances of derivative-free reinforcement learning in optimization, exploration and computation. In optimization, the article organizes the aspects into parameter updating and model selection. In exploration, the article presents the recently proposed derivative-free exploration methods in reinforcement learning. In computation, the article reviews the parallel and distributed derivative-free reinforcement learning approaches. We also discuss some limitations and potential future directions of derivative-free reinforcement learning.

The rest of this article is organized as follows. In Section 2, the article presents the background and organizes the aspects into introducing the basic concepts of reinforcement learning as well as derivative-free optimization. In Section 3, the article presents the relationship between reinforcement learning and derivative-free optimization from the aspects of optimization, exploration, and computation. This section also discusses why they should be considered together and how to combine them. The recent progress in derivative-free reinforcement learning from the aspect of optimization is reviewed in Section 4 and 5. Specifically, in Section 4, the derivative-free model parameter updating in reinforcement learning is reviewed according to the different optimization methods. And in Section 5, the article presents the derivative-free model selection in reinforcement learning. In Section 6, the derivative-free exploration in reinforcement learning is reviewed. In Section 7, the article presents the parallel and distributed derivative-free reinforcement learning. At last, in Section 8, we conclude the paper, and discuss some current limitations and possible future concerns of this direction.

\section{Background}
This section introduces the background of reinforcement learning and derivative-free optimization. The fusion of them could be an effective way of tackling the hard policy search problems in reinforcement learning.

\subsection{Reinforcement learning}
\begin{figure}[!htbp]
	\centering\includegraphics[width=0.50\textwidth]{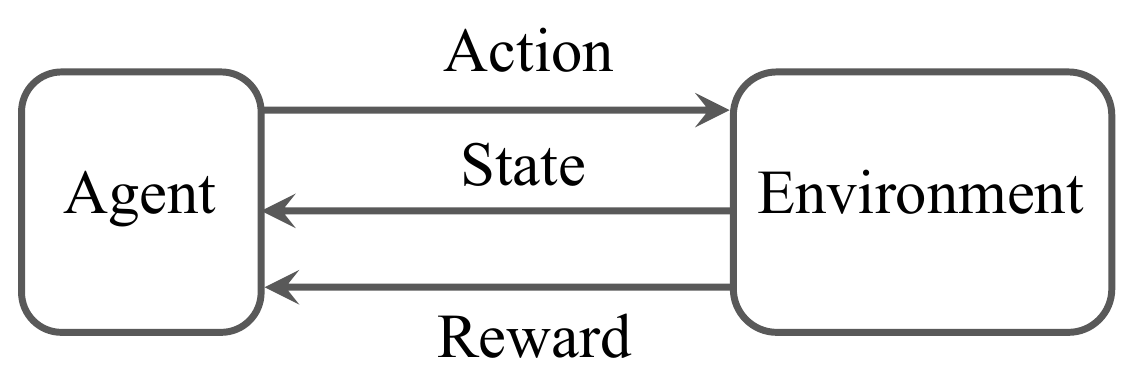}
	\caption{Interaction structure of reinforcement learning.}\label{RL}
\end{figure}

Reinforcement learning (RL)~\cite{sutton_reinforcement_1998,RL-SOTA-book} is an important direction in machine learning~\cite{dietterich_machine_1997}. It aims to enable an agent to automatically learn the policy with the maximum long-term reward via interactions with an environment. An illustration of interaction structure of reinforcement learning is shown in Figure~\ref{RL}. 
When an agent is put in an unknown environment, it is told of the action space $A$ in which there are actions it can choose to take, and the state space $S$ in which its observation is contained in. Both $S$ and $A$ can be discrete or continuous. The agent has a policy $\pi$ to determine which action $a\in A$ is chosen at a state $s\in S$. Generally, a deterministic policy $\pi$ is a mapping from the state space $S$ to the action space $A$, i.e., $a=\pi(s)$; while a stochastic policy $\pi$ chooses the action $a$ according to the probability distribution $\pi(a|s)$ at the state $s$ with the constraints 
\begin{equation}
\forall a\in A,\ \pi(a|s)\geq 0\quad\text{and}\quad\sum_{a\in A}\pi(a|s)=1.
\end{equation}
When the agent takes one action $a_t$ at a state $s_t$ in time step $t$, the unknown environment typically responds by transiting to the next state $s_{t+1}$ and feeds back a reward signal. The (unknown) transition function can be represented as a distribution $P(s_{t+1}|s_t,a_t)$, and the (unknown) reward function can be represented as $r_t=R(s_t,a_t)$.

The agent explores the environment according to a policy $\pi$ by interactions with the environment. Starting from the initial state $s_0 \sim \rho_0(\cdot)$, where $\rho_0(\cdot)$ is the start state distribution, the interaction trajectory $\tau$ (also frequently termed as episode or roll-out) is
\begin{equation}
\begin{aligned}
& s_0, a_0\sim\pi(\cdot|s_0),s_1\sim P(\cdot|s_0,a_0),r_0=R(s_0,a_0),\\
& a_1\sim\pi(\cdot|s_1), \ldots
\end{aligned}
\end{equation}
The quality of the policy $\pi$ is then evaluated from interaction trajectories, as the expected sum up of the reward along the trajectories. Commonly, the expected total reward, or expected return, of the policy $\pi$ starting from the state $s$ is the expectation over the policy distribution and the transition distribution, e.g., 
\begin{equation}
V^{\pi}(s)=\mathbb{E}\big[\sum_{t=0}^{+\infty}\gamma^{t}r_t|s_0=s\big],
\end{equation}
where $\gamma \in [0,1)$ is the discount factor.

Note that reinforcement learning setting shares the key components $<A,S,P,R,\gamma>$ with Markov decision process (MDP)~\cite{bellman1957markovian}, except that all the functions are known under the definition of an MDP. Also note that dynamic programming is a classical and effective method to solve the best policy in MDPs. There are thus two branches of approaches, model-based ones that recover the transition distribution and reward function to form up a complete MDP for using classical solvers, and model-free ones that learn the policy without recovering the MDP. In this paper, we mainly focus on model-free approaches. Nevertheless, model-assisted model-free reinforcement learning is a promising direction.

In model-free reinforcement learning, the policy can be derived from well learned value functions, or can be learned directly. In either ways, the general steps of the learning approaches are the same, i.e., iterating between exploring the environment and updating the policy. The exploration is typically implemented by executing the policy with added noise, such as $\epsilon$-greedy and Gibbs sampling. In this way, the probability of executing every action and visiting every state is non-zero, and thus diverse trajectory data can be generated for the next step of policy update.

To learn the value function, the most popular way might be the temporal difference update~\cite{sutton_reinforcement_1998}, which is essentially an incremental update rule of arithmetic sum
\begin{equation}
V^{\pi}(s_t) \leftarrow V^{\pi}(s_t)+\delta \cdot  \big(r_t+V^{\pi}(s_{t+1})-V^{\pi}(s_t)\big),
\end{equation}
where $\delta>0$ denotes the step size. After obtaining the value function, the policy is derived simply as that taking the action of the largest value, i.e., $\pi(s)=\argmax_{a}Q^{\pi}(s,a)$, where $Q^{\pi}(s,a)=R(s,a)+\mathbb{E}_{s'\sim P(\cdot|s,a)}V^{\pi}(s')$. $V^{\pi}(s)$ and $Q^{\pi}(s,a)$ are called value functions.

The value function based methods may face the problem of policy degradation~\cite{Bartlett_2001}. That is to say, a more accurate estimation of the value function may not guarantee a better policy. Another kind of approach is to learn the policy directly, i.e., policy search. In policy search, the policy model is firstly parameterized, such as softmax policy for discrete action space
\begin{equation}
\pi_{\theta}(a|s)=\frac{\exp\big(h_{\theta}(s,a)\big)}{\sum_{a'\in A}\exp\big(h_{\theta}(s,a')\big)}, \label{eq-1}
\end{equation}
and Gaussian policy for continuous action space
\begin{equation}
\pi_{\theta}(a|s) = \frac{1}{\sqrt{2\pi}\sigma}\exp\left(-\frac{(a-h_{\theta}(s))^2}{\sigma^2}\right), \label{eq-2}
\end{equation}
where $h_{\theta}(s,a)$ in Eq.~(\ref{eq-1}) and $h_{\theta}(s)$ in Eq.~(\ref{eq-2}) with parameter $\theta$ can be linear, and are nowadays often (deep) neural network models. Note that this parametric policy model is fully differentiable with respect to (w.r.t.) $\theta$, if $h_{\theta}$ is differentiable. Then, consider the objective of learning an optimal policy $\pi^* = \pi_{\theta^*}$ that can maximize the total reward $J(\theta)$, i.e., 
\begin{equation}
\theta^* = \argmax_{\theta\in\Theta} J(\theta), \label{eq-3}
\end{equation}
where $\Theta$ denotes the parameter space. We suppose that both the environment transitions and the policy are stochastic. In episodic environments, $J(\theta)$ can be trajectory-wise total reward
\begin{equation}
J(\theta) = \mathbb{E}_{\tau\sim\pi_{\theta}}[R(\tau)] = \int_{\mathcal{T}} p_{\theta}(\tau)R(\tau)\,\mathrm{d}\tau, \label{eq-4}
\end{equation}
where $R(\tau)=\sum_{t=0}^{T-1} r_t$ is the total reward (or return) over a trajectory $\tau$, $\mathcal{T}$ is a valid trajectory space, and $p_{\theta}(\tau) = \rho_0 (s_0) \prod_{t=0}^{T-1} P(s_{t+1}|s_t, a_t) \pi_{\theta}(a_t|s_t)$ is the probability of generating a $T$-step trajectory $\tau$ according to $\pi_{\theta}$ in the environment. In continuing environments, $J(\theta)$ can be average reward per time-step for one-step MDPs
\begin{equation}
J(\theta)=\int_S d^{\pi_{\theta}}(s)\int_A \pi_{\theta}(a|s)R(s,a)\, \mathrm{d}a\,\mathrm{d}s, \label{eq-5}
\end{equation}
where $d^{\pi_{\theta}}(s)$ is the stationary probability of visiting the state $s$ following the policy $\pi_{\theta}$. We can find that the total reward objective $J(\theta)$ is also differentiable w.r.t. $\theta$. A straightforward idea of solving the parameter is to follow the gradient $\nabla_{\theta}J(\theta)$, which is generally called as policy gradient method.

Affected by the deep learning, flexible and capable deep neural network models are introduced in reinforcement learning. It can be noticed that, although the objective $J(\theta)$ is differentiable, the objective function is not simple, particularly when deep neural network models are employed. How to best optimize the objective function is still an open problem with continual progress.

\subsection{Derivative-free optimization}
Optimization, $x^* = \argmax_{x\in \mathcal{X}}f(x)$ as a general representative, plays a fundamental role in machine learning. With the rapid development of machine learning, it has deeper applications in a broader and more complex scenario. Due to the increasing complexity of machine learning application problems, optimization methods for complex and hard problems are attracting more and more attention from researchers. For complex optimization problems (e.g., non-convex, non-smooth, non-differential, discontinuous, and NP-hard, etc.), gradient-based methods may fall into the local optima or even lose their power, and result in the unsatisfactory performance.

In a search space, the objective of optimization is to find the solution with the extreme function value. A general principle of optimization is simply that, given the currently accessed solutions, which can be randomly sampled at the initialization, find the next solution with better function value. For example, gradient ascent method finds the next solution follows the gradient direction of the objective function. This procedure requires firstly that the objective function is differentiable, and more importantly that there are few local optima and saddle points, so that the gradient direction is informative and can lead to better solutions. 

Derivative-free optimization (DFO)~\cite{DFO-book,DFO-siamrev,DFO-review}, also termed as zeroth-order or black-box optimization, finds the next solutions in another way. It covers many families of optimization algorithms that do not rely on the gradient. It only relies on the function values (or fitness values) $f(x)$ on the sampled solution $x$. Most derivative-free optimization algorithms share a common structure. They firstly initialize from some random solutions in the search space. From the accessed solutions, derivative-free optimization algorithms build a model, either explicitly or implicitly, about the underlying objective function. The model could imply an area that contains some potential better solutions. They then sample new solutions from that model and update the model. Derivative-free optimization methods repeat this sampling-and-updating procedure to iteratively improve the quality of solutions. To sum up, the general framework of derivative-free optimization methods involve some key steps below:
\begin{enumerate}
	\item Randomly sample solutions;
	\item Evaluate the objective function value of the sampled solutions;
	\item Update the model from the sampled solutions and their function values;
	\item Sample new solutions according to the model with a designed mechanism;
	\item Repeat from step 2 until some termination conditions are satisfied;
	\item Return the best found solution and its function value.
\end{enumerate}
The termination conditions usually include that the function evaluation budget is exhausted or the goal optimal function value has been reached. 
Representative derivative-free algorithms include evolutionary algorithms~\cite{Holland-1975,HansenMK03}, Bayesian optimization~\cite{reviewBO16}, cross-entropy method~\cite{ce-BoerKMR05}, deterministic or stochastic optimistic optimization~\cite{MunosFTML2014}, and classification-based optimization~\cite{yu.qian.racos}, etc. So far, the substantial progress has been made in both the theoretical analysis tools and theoretical guarantees of derivative-free optimization~\cite{AIJ-HeYao01,Yu08aij,jmlr-Bull11,nips-JamiesonNR12,MunosFTML2014,cec-YuQian14,tit-DuchiJWW15,tec-YuQZ15,Kawaguchi15nips,yu.qian.racos,jair-KawaguchiMZ16}. Some fundamental and crucial concerns in theory, such as the global convergence rate and the function class that can be optimized efficiently, are disclosed progressively.

In order to take a deeper insight of the key step 3 and step 4 in the above procedure, we present a snapshot of a canonical genetic algorithm~\cite{Holland-1975,books-GA98} as an example. Genetic algorithms belong to the family of evolutionary algorithms. Consider maximizing a pseudo-Boolean function $\argmax_{x\in\{0,1\}^d}f(x)$. Here the solutions are represented as bit strings of length $d$. A canonical genetic algorithm maintains a population which includes $n$ solutions, and the model in each iteration/generation is represented by the population (i.e., $n$ solutions in each population). In each iteration, $n$ children (or offspring) are produced on the basis of $n$ parent solutions via the designed mechanism called crossover operator from the parents, e.g., single-point crossover which is shown below. We set $d=5$ for illustration.
\begin{equation}
\begin{aligned}
x^{\text{parent}}_1 = (0,\underline{1},0,0,1)\\
x^{\text{parent}}_2 = (1,\underline{0},1,0,0)
\end{aligned}
\xrightarrow{\text{crossover}}
%\stackrel{\text{crossover}}\Longrightarrow
\begin{aligned}
x^{\text{child}}_1=(0,\underline{1},1,0,0)\\
x^{\text{child}}_2=(1,\underline{0},0,0,1)
\end{aligned}
\end{equation}
In the above single-point crossover, a point (or bit) on both parents' bit strings is picked randomly, and bits to the right of that point are swapped between the two parents, which results in two children. Besides, mutation is another variation operator in the designed mechanism. If mutation takes place, it can be applied to each child that has been produced by crossover. One way of realizing mutation could be flipping each bit in a solution $x^{\text{child}}$ with some probability and producing a new solution $x^{\text{child}}_{\text{new}}$, which is shown below. We set $d=5$ for illustration.
\begin{equation}
x^{\text{child}}=(0,1,1,\underline{0},0) 
\xrightarrow{\text{mutation}}
%\Longrightarrow 
x^{\text{child}}_{\text{new}}=(0,1,1,\underline{1},0)
\end{equation}
After the crossover and mutation operators (step 4), the objective function values of these newly produced solutions are evaluated (step 2). Then, the model, which is represented by the solutions, is updated via selecting the best $n$ solutions from both the parents and children to form the next generation of population (step 3).

From the above algorithmic procedure, some key characteristics of derivative-free optimization algorithms can be observed. Firstly, they can be utilized as long as the quality of solutions can be evaluated. Secondly, the designed mechanisms for sampling solutions and rules for updating model always consider the balance between exploration and exploitation. In optimization, exploration intuitively means gathering more information about objective functions and reducing some uncertainty, while exploitation means choosing the best solutions under current information. In the instance of canonical genetic algorithm, exploration is realized via crossover and mutation operators. The combination of exploration and exploitation could help optimization procedures maintain global and local search, and jump out of the local optima in order to find out the (approximately) global optima with high probability. Thirdly, many derivative-free optimization algorithms are population-based. A population of solutions is maintained and improved iteratively, and the algorithms could share and leverage the information across a population. These key characteristics make derivative-free optimization methods have a low barrier to use as well as the ability of search globally.

Since the conditions of using derivative-free optimization methods are relaxed, they are easy to use and general in the continuous, discrete, or mixed search space. Their ability could guarantee the effectiveness of global optimization. Therefore, derivative-free optimization methods are suitable for dealing with the complex and hard optimization problems. They have been applied in the complex learning tasks and achieved impressive empirical results, such as policy search in reinforcement learning~\cite{TaylorWS06,AbdolmalekiLPLR15,hu2017sequential,salimans2017evolution}, automatic machine learning and hyper-parameter tuning~\cite{SnoekLA-nips12,Thornton13kdd,Real17icml,Real18}, objective detection in computer vision~\cite{cvpr-ZhangSVPL15}, subset selection~\cite{qianc15-nips,qianc17-ijcai}, and security games~\cite{BrownAKOT14}, etc.

\section{When RL meets DFO}
After a brief recap of reinforcement learning (RL) and derivative-free optimization (DFO), this section explains and emphasizes that the fusion of them, termed as \emph{derivative-free reinforcement learning}, is necessary and attractive. We explain it from the aspects of optimization, exploration, and computation, respectively.

\textbf{Optimization.} A learning task typically involves three components, and they are representation, evaluation, and optimization~\cite{cacm-Domingos12}. The ability of optimization method has a significant impact on the complexity of model representation and the type of evaluation function that we can choose for learning. Nowadays, in reinforcement learning, the policy or value function models are often represented by deep neural networks, i.e., deep reinforcement learning. When injecting this sophisticated deep representation into the evaluation function in RL (e.g., Eq.~(\ref{eq-4}) or Eq.~(\ref{eq-5}) in Section~2.1), the resulting optimization problems (e.g., Eq.~(\ref{eq-3}) in Section~2.1) could be non-convex.

For such optimization problems, standard gradient-based methods may suffer from the difficulties of stationary point issues (e.g., a plethora of saddle points or spurious local optima), bad condition number, or flatness in the activations that could lead to the gradient vanishing problem~\cite{shalev-shwartz_failures_2017}. And these difficulties could result in the unsatisfactory results.

At the same time, derivative-free methods that conduct optimization from samples provide another way of policy learning, and can be complementary with gradient-based ones in reinforcement learning. One straightforward way of applying derivative-free optimization methods is to define the search space as the policy parameter space and the objective function as the expected long-term reward. Namely, $\mathcal{X} \overset{\text{def}}{=} \Theta$ and $f(x) \overset{\text{def}}{=} J(\theta)$. For policy learning, derivative-free optimization methods have their own merits of being able to search parameters globally and being easy to train. They do not perform gradient back-propagation, do not care whether rewards are sparse or dense, do not care the length of time horizons, and do not need value function approximation~\cite{salimans2017evolution}. In Section~4 and 5, we will discuss the recent advances dedicated to the derivative-free model parameter updating as well as model selection in reinforcement learning, respectively.

\textbf{Exploration.} Almost all reinforcement learning methods share the exploration-learning framework~\cite{Yu18-ijcai}. Namely, an agent explores and interacts with an unknown environment to learn the optimal policy that maximizes the total reward from the exploration samples. Generally speaking, the exploration samples involve states, state transitions, actions and rewards. From the exploration samples, the quality of a policy can be evaluated by rewards, and the learning step updates the policy or value function models from the evaluations. This exploration-learning procedure is repeated until some termination conditions are met. Exploration is necessary in reinforcement learning. Because achieving the best total reward on the current trajectory samples is not the ultimate goal, and the agent should visit the states that have not been visited before so as to collect better trajectory samples. This means that the agent should not follow the current policy tightly, and thus the exploration mechanisms need to encourage the agent to deviate from the previous trajectory paths properly. 

Most existing exploration mechanism mainly suffer from being memoryless and blind, e.g., action space noise or parameter space noise~\cite{PlappertHDSC0AA-iclr18}, or being difficult to use in real state/action spaces, e.g., curiosity-driven exploration~\cite{PathakAED-icml17}. On the other hand, many mainstream policy gradient methods, such as truncated natural policy gradient (TNPG)~\cite{DuanCHSA-icml16} and trust region policy optimization (TRPO)~\cite{SchulmanLAJM-icml15}, seldom touch the exploration.

Meanwhile, derivative-free reinforcement learning is naturally equipped with the exploration strategies. Because in the search process of derivative-free optimization methods, the designed mechanisms for sampling solutions and rules for updating model always consider the exploration. Therefore, derivative-free optimization methods can take part of the duty of exploration for reinforcement learning when updating the policy or value function models from samples. Recently, some problem-dependent derivative-free exploration methods that could improve the sample efficiency have been proposed. In Section~6, we will discuss these works dedicated to the derivative-free exploration in reinforcement learning.

\textbf{Computation.} Although derivative-free methods could bring some good news to reinforcement learning with respect to optimization and exploration, they mostly suffer from low convergence rate. Derivative-free optimization methods often require to sample a large amount of solutions before convergence, even if the objective function is convex or smooth~\cite{nips-JamiesonNR12,tit-DuchiJWW15,BachP16-colt16}. And the issue of slow convergence becomes more serious as the dimensionality of a search space increases~\cite{tit-DuchiJWW15,aaai-QianYu16}. Obviously, this issue will block the further application of derivative-free methods to reinforcement learning. They sample a lot of policy parameters before finding out an optimal or satisfactory one, and the quality of policy parameters is evaluated by the trajectory samples. This makes reinforcement learning more sample inefficient.

Fortunately, many derivative-free optimization methods are population-based. That is to say, a population of solutions is maintained and improved iteratively. This characteristic makes them highly parallel. Thus, derivative-free optimization methods can be accelerated by parallel or distributed computation, which alleviates their slow convergence. Furthermore, for parallel/distributed derivative-free methods, the data communication cost is lower compared with gradient-based ones, since only scalars (fitness values) instead of gradient vectors or Hessian matrices need to be conveyed. This merit further compensates for the low convergence rate partly. In Section~7, we will discuss the recent works dedicated to the parallel/distributed derivative-free reinforcement learning.

\begin{figure}[!htbp]
	\centering\includegraphics[width=0.68\textwidth]{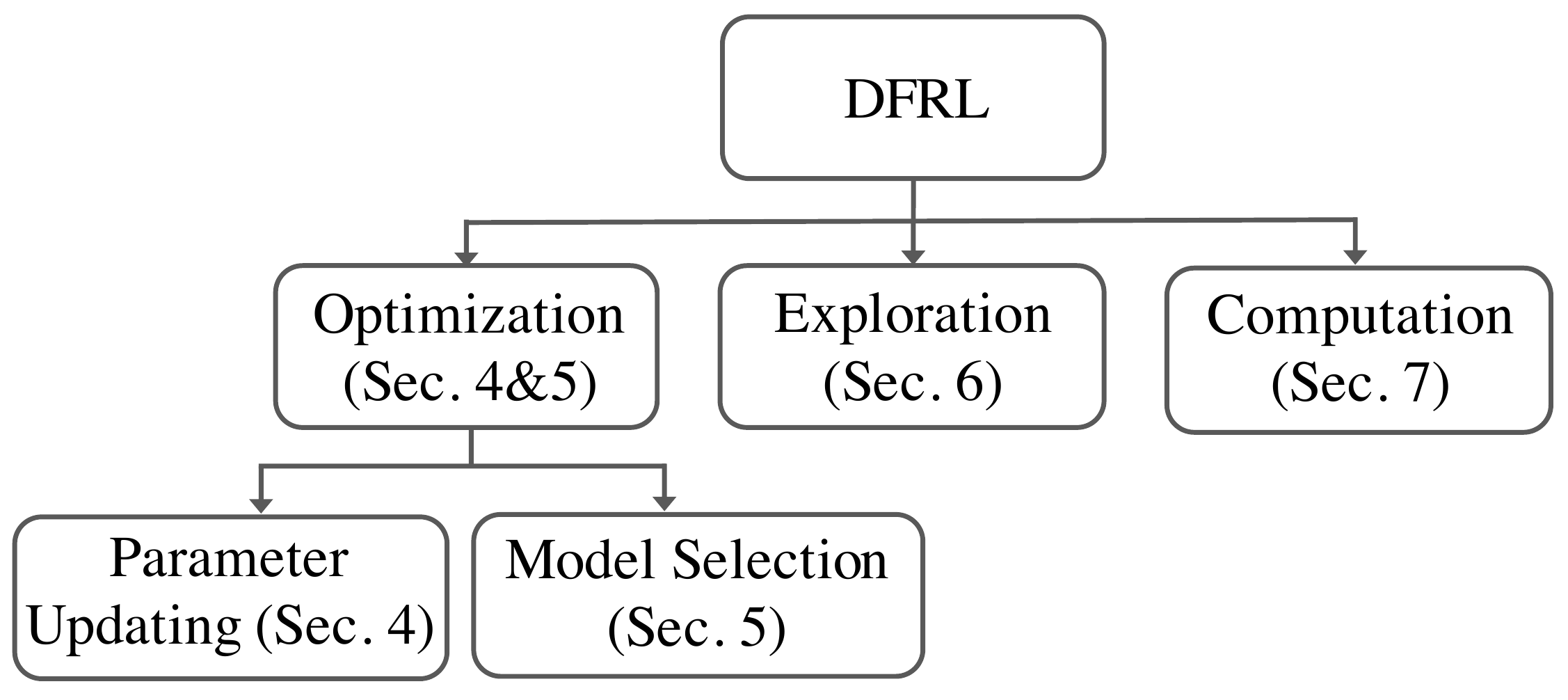}
	\caption{The organization of the works reviewed in the article.}\label{structure}
\end{figure}

To sum up, derivative-free reinforcement learning could hopefully result in more effective and powerful algorithms for complex control tasks, more sample efficient exploration in environments, and more time efficient global optimization for better policy. The organization of the following reviewed works is depicted in Figure~\ref{structure}. We should stress that derivative-free optimization methods are not proposed to replace gradient-based ones, e.g, policy gradient algorithms, in reinforcement learning, and they are complementary with each other. In fact, some impressive works that will be discussed in this article are hybrids of derivative-free and gradient-based ones.

\section{Derivative-free model parameter updating in reinforcement learning}
The generic framework of using derivative-free optimization algorithms to update the model parameters in reinforcement learning is quite straightforward. After parameterizing the policy or value function models, the quality of a parameter $\theta$ is evaluated via the total reward $J(\theta)$ provided by an environment. For policy search methods in deep reinforcement learning, the parameters $\theta$ of a policy model $\pi_{\theta}$ are the weights of a deep neural network. Here the policy model can be softmax policy for discrete action space or Gaussian policy for continuous action space, and this section only considers that the architecture/topology of a (deep) neural network is fixed. Derivative-free algorithms regard $\theta$ and $J(\theta)$ as solutions and objective function values (or fitness values) respectively, and search the optimal solution $\theta^*$. They sample different policy parameters, and learn where to sample in the next iteration.

The works dedicated to applying derivative-free methods to optimize the weights of neural networks have been developed for decades, e.g., neuroevolution. In neuroevolution, evolutionary algorithms, inspired from natural evolution, can not only optimize the weights, but also search the 
topology of neural networks (discussed in Section~5). In each generation of a neuroevolutionary algorithm, each neural network in the population is evaluated by the task, and the best ones are selected. The crossover and mutation operators are used to reproduce the new networks from the selected ones in order to form a new population, and the process iterates. A comprehensive discussion of neuroevolution can be found from the last century review paper~\cite{yao1999evolving} or the recent one~\cite{stanley-neuroevolution-nature19}. 

The methods of applying neuroevolution for reinforcement learning tasks have been developed for decades. Recently, more and more works have shown that, compared with gradient-based methods, neuroevolution is competitive for not only policy search in RL~\cite{salimans2017evolution,such_deep_2017}, but also supervised learning tasks~\cite{MorseS-gecco16,relationship-17}. This confirms the power of neuroevolution and renews the increasing interests in it. In addition to the success of neuroevolution RL~\cite{KoutnikCSG-gecco13,HausknechtLMS-tciaig14,salimans2017evolution,RisiT-tciaig17,chrabaszcz2018back}, other derivative-free optimization methods for RL could also be promising. For instance, stochastic zeroth-order search could rival the gradient-based methods for the static linear policy optimization on the MuJoCo locomotion tasks~\cite{ManiaGR-nips18}. And its convergence behavior is analyzed in~\cite{MalikPBKBW-aistats19}.

In this section, we mainly review the recent progress in derivative-free model parameter updating in reinforcement learning, and the involved works include but are not only restricted to neuroevolution RL. For the early works on evolutionary RL, the survey of them can be found in~\cite{jair-MoriartySG99,Whiteson12}.

\subsection{Evolution strategies based model parameter updating}
Evolution strategies (ESs)~\cite{HansenAA-15} belong to the family of evolutionary algorithms inspired from natural evolution. In ESs, a parameterized search distribution (e.g., Gaussian or Cauchy distribution) is maintained, and solutions are sampled from this search distribution and then are evaluated by an objective function. The search distribution is iteratively updated according to the evaluated solutions. Mutation is a commonly used variation operator in ESs, and it perturbs solutions in order to generate the new ones. Mutation introduces the diversity of solutions in the population, and may alleviate the problem that optimization algorithms are trapped into the local optima. Due to the different algorithmic implementations, ESs have many variants, and one of the most popular among them might be the covariance matrix adaptation evolution strategy (CMA-ES)~\cite{HansenO96,HansenO01,HansenMK03}. In CMA-ES, the search distribution is a multivariate Gaussian, and its covariance matrix adapts over time. The mutation ellipsoids of CMA-ES are not restricted to be axis-parallel, and the overall step size is controlled with the cumulative step size adaptation.

Heidrich-Meisner and Igel~\cite{Heidrich-MeisnerI08,Heidrich-MeisnerI09} proposed to optimize the parameterized policy in RL via CMA-ES. They consider to apply CMA-ES because it is robust (ranking policies based), can detect the correlations among parameters, and can infer the search direction from the scalar signals in RL. In~\cite{Heidrich-MeisnerI08}, CMA-ES is used to optimize linear policies (e.g., $\pi_{\theta}(s)=\theta^{\top}s$ for the deterministic policies) on the double cart-pole balancing task. Empirical results show that, compared with the episodic natural actor-critic algorithm (NAC)~\cite{PetersS08-NAC} which is a gradient-based one, CMA-ES is more robust with respect to noise, policy initialization, and hyper-parameters. On the other hand, NAC could surpass CMA-ES with respect to learning speed under appropriate policy initialization and hyper-parameters. In~\cite{Heidrich-MeisnerI09}, CMA-ES is used to search neural network policies on the Markovian and non-Markovian variants of the pole balancing task. Compared with the policy gradient methods, value function based methods, random search, and several neuroevolution methods, overall, the empirical performance of CMA-ES is superior.

Also, Heidrich-Meisner and Igel~\cite{Heidrich-MeisnerI-icml09} enhanced CMA-ES for direct parameterized policy search with an adaptive uncertainty handling mechanism. On the basis of this mechanism, the resulting Race-CMA-ES can dynamically adapt the number of roll-outs for evaluating the parameterized policies such that the ranking of solutions is just reliable enough to push the learning process. Empirical results on the mountain car and swimmer control tasks with linear policies show that, compared with CMA-ES and NAC~\cite{PetersS08-NAC}, Race-CMA-ES can accelerate the learning process and improve the algorithmic robustness. 
Stulp and Sigaud~\cite{StulpS-icml12} were the first to make the relationship between CMA-ES and the cross-entropy (CE) method~\cite{ce-BoerKMR05} explicit under the view of probability-weighted averaging. The efficacy of CE for playing Tetris has been studied in~\cite{SzitaL-neco06}. They claim that CE can be regarded as a special case of CMA-ES through setting some CMA-ES parameters to extreme values. Besides, they also inject the covariance matrix adaptation into the policy improvement with path integrals, and result in PI$^\text{2}$-CMA. PI$^\text{2}$-CMA shares the similar way of performing the parameter updating with CMA-ES and CE, but has the more consistent convergence behavior under varying initial conditions.

Wierstra et al.~\cite{WierstraSPS-cec08,WierstraSGSPS-jmlr14} presented the natural evolution strategies (NESs) where the natural gradient was used to update a parameterized search distribution in the direction of higher expected fitness. The effectiveness of NESs is empirically verified on a neuroevolutionary control policy design for the non-Markovian double pole balancing task. And the result shows that NESs possess the merits of alleviating oscillations and premature convergence.

Salimans et al.~\cite{salimans2017evolution} from OpenAI proposed to use a simple evolution strategy, based on a simplification of NESs, to directly optimize the weights $\theta$ of policy neural networks. After initializing the policy parameters $\theta_0$, OpenAI ES stochastically mutates the parameters $\theta$ of the policy with Gaussian distribution, and evaluates the resulting mutated parameters via the total reward/return $J(\cdot)$. Then, it combines these returns and updates the parameters $\theta$. This process iterates till the termination condition is met. Let $\alpha>0$ denote the learning rate, and $\sigma$ denote the noise standard deviation. The core steps are summarized as follows:
\begin{enumerate}
	\item Randomly sample $\epsilon_1,\ldots,\epsilon_n$ from the standard Gaussian distribution $\mathcal{N}(0,I)$;
	\item Compute returns $J_i = J(\theta_t+\sigma\epsilon_i)$ for $i=1,\ldots,n$;
	\item Update parameters $\theta_{t+1}\leftarrow \theta_t+\alpha\frac{1}{n\sigma}\sum_{i=1}^{n}J_i\epsilon_i$;
	\item Repeat from step 1 till the termination condition is met.
\end{enumerate}
Furthermore, some techniques are integrated to facilitate the success of OpenAI ES for learning policy neural networks. To name a few, virtual batch normalization~\cite{SalimansGZCRCC-nips16} is adopted to enhance the exploration ability, antithetic sampling~\cite{GewekeAntithetic-1988} (also known as mirrored sampling~\cite{BrockhoffAHAH-ppsn10}) is adopted to reduce variance, and fitness shaping~\cite{WierstraSGSPS-jmlr14} is adopted to decrease the trend of trapping into the local optima in the early training phase. This simple and generic OpenAI ES is very easy to parallelize and only takes low communication cost, which will be discussed in Section~7. The effectiveness of OpenAI ES is empirically investigated on some simulated robotics tasks in MuJoCo and Atari games in OpenAI Gym~\cite{MuJoCo-12,Atari2600-jair13,openai-gym}. And the results surprisingly show that OpenAI ES is comparable to some state-of-the-art gradient-based algorithms, e.g., trust region policy optimization (TRPO)~\cite{SchulmanLAJM-icml15} and asynchronous advantage actor-critic (A3C)~\cite{mnih_asynchronous_2016}, on both simple and hard environments, but OpenAI ES needs more samples. Notably, this work~\cite{salimans2017evolution} seems to renew the increasing interests in (deep) neuroevolution RL.

Encouraged and inspired by~\cite{salimans2017evolution}, new ideas, discussions and approaches are blooming out. 
Lehman et al.~\cite{LehmanCCS-gecco18a} claimed that the simple OpenAI ES in~\cite{salimans2017evolution} was more than just a finite-difference approximation of the reward gradient. Because OpenAI ES optimizes the average return of the whole solutions in the population instead of a single solution, it searches parameters that are robust to perturbation in the parameter space and yields more stable policies.
Chrabaszcz et al.~\cite{chrabaszcz2018back} compared a simper and basic canonical ES algorithm with OpenAI ES~\cite{salimans2017evolution}. The empirical results on a subset of 8 Atari games~\cite{Atari2600-jair13,openai-gym} show that this simpler canonical ES is able to match or even surpass the performance of OpenAI ES. This work~\cite{chrabaszcz2018back}, on the one hand, further confirms that the power of ES-based model parameter updating for policy search may rival that of the gradient-based algorithms. On the other hand, it indicates that the ES-based RL algorithms could be further ameliorated in many aspects by integrating the recent advances made in the field of ES.

Choromanski et al.~\cite{ChoromanskiRSTW-icml18} enhanced ES-based RL model parameter updating by structured evolution and compact policy networks. The resulting algorithm needs less policy neural network parameters, and thus could speed up training and inference. 
Chen et al.~\cite{ChenZHJ-ijcai19} improved ES for training policy neural networks in terms of the mutation strength adaptation and principal search direction update rules. And the number of elitists adaptation and restart procedure are also integrated to tackle the local optima. The proposed algorithm is of linear time complexity and low space complexity, and thus possesses the merit of scalability. 
Choromanski et al.~\cite{Choromanski-nips19} applied the ideas of active subspaces~\cite{activeSubspace-SIAM15}, which is one of the popular approaches to dimensionality reduction, to yield the sample-efficient and scalable ES for policy optimization in RL. 
Liu et al.~\cite{LiuZYBQYL-aaai19} boosted the sample efficiency of ES for RL by reusing the existing sampled data, and proposed the trust region evolution strategies (TRES) algorithm. TRES realizes sample reuse for multiple epochs of updates in the way of iteratively optimizing a surrogate objective function. 
Tang et al.~\cite{vr4es} proposed a variance reduction technique for ES-based RL model parameter updating. It utilizes the underlying MDP structure of RL through problem re-parameterization and control variable construction. 
Fuks et al.~\cite{FuksAHL-ijcai19} proposed a technique called progressive episode lengths (PEL), and injected it into a canonical ES~\cite{chrabaszcz2018back} to result in PEL-ES. Inspired from transfer learning and curriculum learning, PEL-ES firstly lets an agent play the easy and short tasks, and then applies the gathered knowledge to further deal with the harder and longer tasks. Compared with the canonical ES~\cite{chrabaszcz2018back}, PEL-ES is superior in optimization speed, total rewards, and stability.

Houthooft et al.~\cite{HouthooftCISWHA-nips18} proposed a meta-learning method for policy learning across different tasks called evolved policy gradient (EPG). EPG encodes the prior knowledge implicitly through a parametrized loss function, and parametrization is realized by temporal convolutions over the agent's experience. The agent can use the learned loss function to learn on a new task quickly. Thus, the main procedure is to evolve a parametrized and differentiable loss function. The agent optimizes its policy to minimize this loss function so as to gain high returns. To implement this procedure, EPG involves two optimization loops:
\begin{itemize}
	\item In the inner loop, the agent learns to solve a task through minimizing a loss function provided by the outer loop.
	\item In the outer loop, the parameters of a loss function are optimized in order to maximize the returns gained after the inner loop.
\end{itemize}
\begin{figure}[!htbp]
	\centering\includegraphics[width=0.50\textwidth]{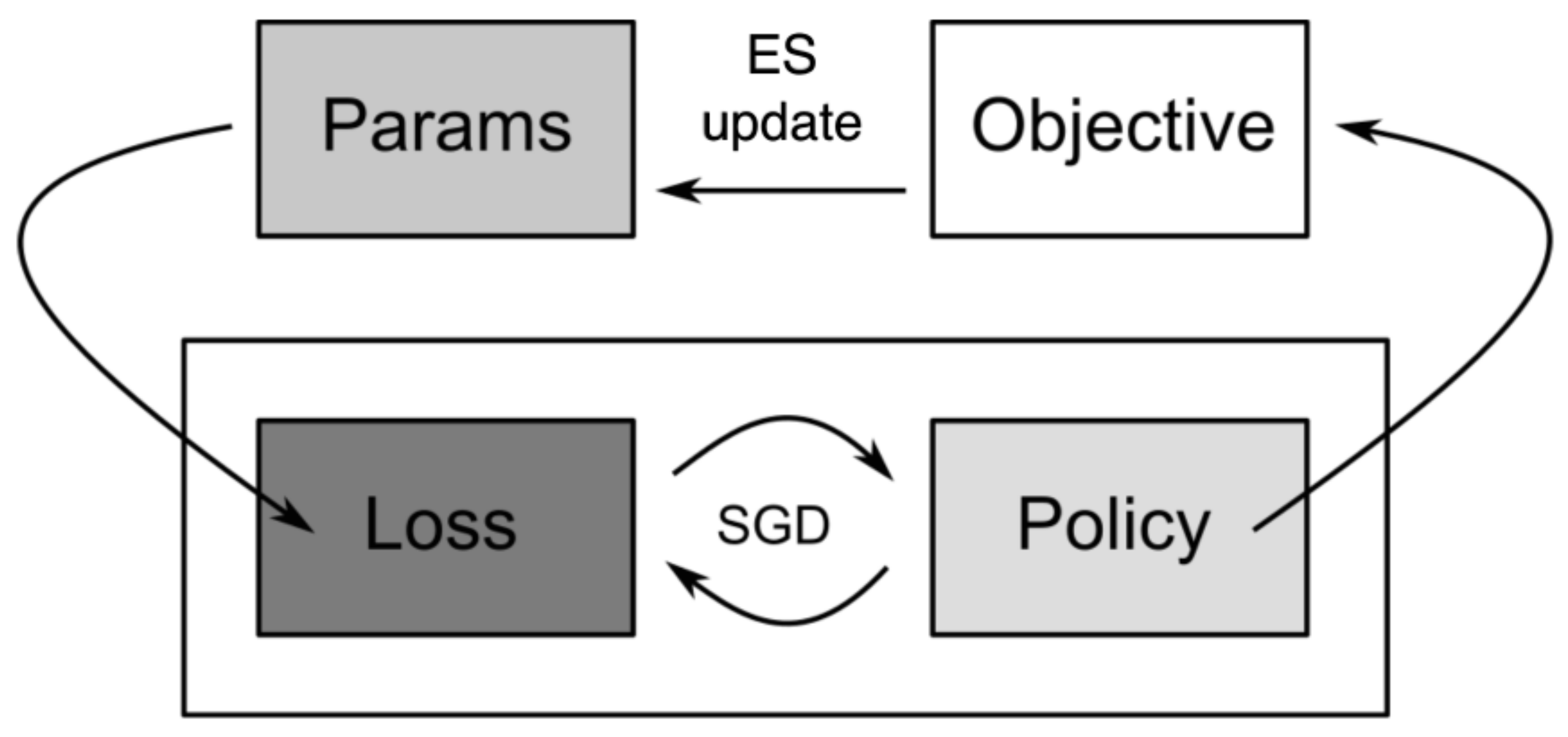}
	\caption{Figure from~\cite{HouthooftCISWHA-nips18} that illustrates the high-level procedure of EPG.}\label{epg}
\end{figure}
The inner loop is optimized by the stochastic gradient descent (SGD) method. For the outer loop, since the returns are not explicit functions of the loss function parameters, ES can be applied to optimize the parameters of this loss function. Figure~\ref{epg} illustrates the main procedure of EPG. Empirical results on several continuous control tasks in MuJoCo~\cite{MuJoCo-12,openai-gym} show that, compared with proximal policy optimization (PPO)~\cite{SchulmanWDRK17-PPO} that is an off-the-shelf policy gradient algorithm, EPG can generate a learned loss function which trains the agent faster. Besides, EPG also exhibits generalization properties for out-of-distribution test time tasks that surpass other meta-learning algorithms RL$^\text{2}$~\cite{DuanSCBSA16} and MAML~\cite{FinnAL-icml17}. Notably, in EPG, ES is used to optimize the parameters of a loss function instead of policy parameters. And the success of EPG largely owes to the hybrid of derivative-free and gradient-based algorithms. The similar scenario also happens in~\cite{HaS-nips18,YuLT-iclr19}, where CMA-ES and gradient-based algorithms are mixed for policy transfer from simulation to reality.

\subsection{Genetic algorithms based model parameter updating}
Similar to evolution strategies (ESs), genetic algorithms (GAs)~\cite{books-GA98} also belong to the family of evolutionary algorithms inspired from natural evolution. Genetic algorithms maintain a population of solutions. Via the operators of mutation, crossover (also called recombination) and selection, a population of solutions can be evolved. One of the main differences between GAs and ESs is that the crossover operator is often used in GAs. Crossover combines two parents to generate new offspring, and the newly generated solutions are usually mutated before adding them to the population. Notably, the crossover operator in GAs could further improve the diversity of solutions in the population.

To verify whether GAs can also be effectively applied to the RL tasks, Such et al.~\cite{such_deep_2017} used a very simple GA to update the weights of deep policy neural networks. It iteratively maintains a population of parameter vectors $\theta$, and the mutation operator is realized by the additive Gaussian noise to $\theta$. The elitism technique is employed during evolution. For simplicity, it does not consider crossover. Due to the huge amount of parameters in deep neural networks and the relatively unsatisfied scalability of GA, the work proposes an approach to storing large parameter vectors $\theta$ compactly via representing each $\theta$ as an initialization seed plus the list of random seeds. This compact representation approach is reversible (i.e., each parameter vector $\theta$ can be reconstructed from it), and substantially improves the scalability and efficiency of deep GA. The policy neural networks with more than four million free parameters can be successfully evolved. Besides, novelty search (NS)~\cite{lehman_abandoning_2011} is integrated into this deep GA to avoid the local optima and encourage exploration. And a distributed version of the deep GA on many CPUs across many machines is implemented for acceleration. The experimental results on the Atari and MuJoCo humanoid locomotion tasks~\cite{Atari2600-jair13,MuJoCo-12,openai-gym} indicate that the deep GA is effective and competitive. Overall, it performs roughly as well as DQN~\cite{mnih_human-level_2015}, A3C~\cite{mnih_asynchronous_2016}, and OpenAI ES~\cite{salimans2017evolution}.

\begin{figure}[!tbp]
	\centering\includegraphics[width=0.50\textwidth]{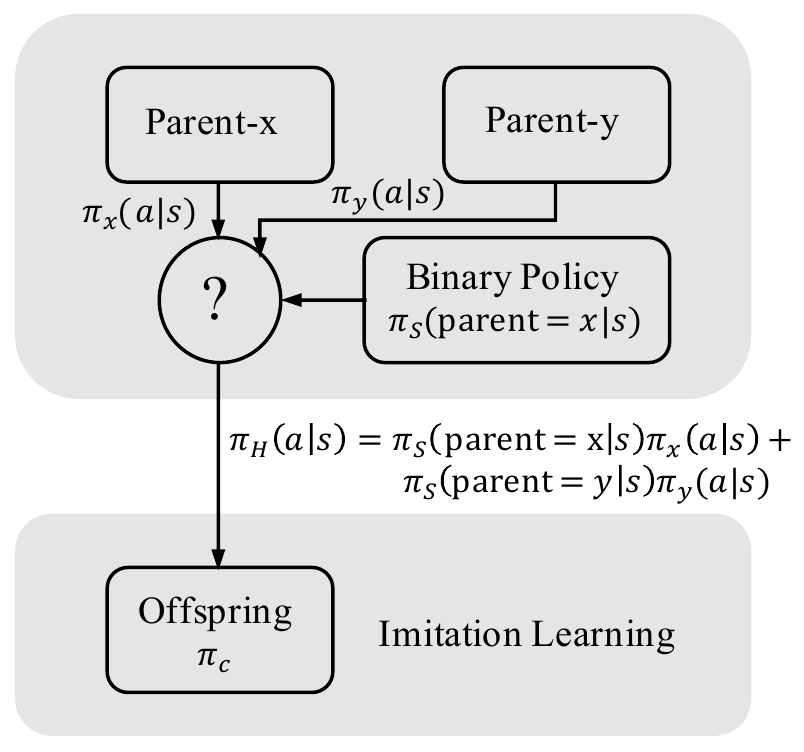}
	\caption{An illustration of combining the parent policies to produce an offspring policy in GPO~\cite{gangwani2018policy}.}\label{parents}
\end{figure}

Gangwani and Peng~\cite{gangwani2018policy} proposed a genetic policy optimization (GPO) method for sample-efficient deep policy optimization. GPO involves crossover, mutation, and selection. For crossover, instead of directly exchanging the parameter representations of two parents (parameter crossover) that may destroy the hierarchical relationship of the neural networks and lead to a catastrophic drop in performance, GPO employs imitation learning to realize policy crossover in the state space. 
The state-space crossover operator combines two parent policies $\pi_x$ and $\pi_y$ to produce an offspring (or child) policy $\pi_c$ that shares the same network architecture as parents via two steps, as shown in Figure~\ref{parents}. Firstly, it trains a two-level policy $\pi_H$, and  
\begin{equation}
\pi_H(a|s) = \pi_S(\text{parent}=x|s)\cdot \pi_x(a|s) + \pi_S(\text{parent}=y|s)\cdot \pi_y(a|s),
\end{equation}
where $\pi_S$ is a binary policy that trains from the parents' trajectories. Given a state $s$, $\pi_H$ selects between the parent policies $\pi_x$ and $\pi_y$, and then outputs the action distribution of the selected parent. Secondly, the trajectories generated from the expert policy $\pi_H$ is used as the supervised data, and the offspring policy $\pi_c$ is trained from the supervised data by imitation learning. To avert the compounding errors introduced by the state distribution mismatch between expert and offspring, the dataset aggregation (DAgger) algorithm~\cite{RossGB11-DAgger} is used for imitation learning. This sample-efficient crossover can distill the knowledge from parents to produce a child that aims to imitate its best parent in generating similar state visiting distributions. 
For mutation, instead of utilizing the commonly-used Gaussian perturbation, GPO mutates the policy network weights by randomly rolling out trajectories and performing several iterations of a policy gradient algorithm using these roll-out trajectories. GPO chooses PPO~\cite{SchulmanWDRK17-PPO} and advantage actor-critic (A2C) algorithm~\cite{sutton_reinforcement_1998,mnih_asynchronous_2016} to fulfill the policy gradient mutation. This mutation operator possesses the merits of high efficiency, sufficient genetic diversity, and good exploration of state space. 
For selection, GPO takes both performance and diversity into consideration. 
The experimental results on some continuous control tasks in MuJoCo~\cite{MuJoCo-12,openai-gym} show that, compared with PPO and A2C which are state-of-the-art policy gradient methods, GPO could be superior in terms of episode reward and sample efficiency. 

Bodnar et al.~\cite{PDERL-aaai20} also made progress in developing the variation operators in GAs tailored to policy optimization. They observe that, when the direct genetic encoding for deep neural networks (i.e., the weights of a network are recorded as a list of real numbers) meets the traditional variation operators in GAs, the negative side-effects may occur in RL since deep neural networks are sensitive to the small variation of weights. For instance, the recently proposed evolutionary reinforcement learning (ERL) framework~\cite{khadka_evolution-guided_2018}, which combines both of them, has the risks of catastrophic forgetting and destructive behaviors, and does not fully address the scalability problem of GAs for RL. Therefore, they propose the learning-based Q-filtered distillation crossover, the safe mutation~\cite{safe-mutation-gecco18} based proximal mutation , and then integrate them into ERL~\cite{khadka_evolution-guided_2018} in a hierarchical manner to result in the proximal distilled evolutionary reinforcement learning (PDERL) framework. PDERL could compensate for the simplicity of the direct genetic encoding, satisfy the functional requirements of the genetic variation operators when applied to the directly encoded deep neural networks, and prevent the catastrophic forgetting of parental behaviors. Compared with PPO~\cite{SchulmanWDRK17-PPO} and twin delayed deep deterministic policy gradients (TD3) algorithm~\cite{TD3-icml18}, as well as ERL, PDERL is superior when evaluated on five robot locomotion tasks in OpenAI Gym~\cite{MuJoCo-12,Atari2600-jair13,openai-gym}. 
Notably, in GPO and PDERL, the gradient-based update is injected into GAs to enhance the genetic variation operators. The success of them implies that, by fusing derivative-free and gradient-based algorithms in a clever way, the strengths of both sides could be absorbed, and more powerful policy search algorithms would be expected.

\subsection{Bayesian optimization based model parameter updating}
Bayesian optimization (BO)~\cite{reviewBO16} is also a kind of widely-used black-box derivative-free approaches aimed at optimizing the difficult functions with relatively few evaluations. BO constructs a probabilistic model for the function being optimized, and then exploits this model to make decisions about the next solution point of the function to evaluate. The newly evaluated solution and the previously evaluated ones are all used to update the probabilistic model so as to improve its accuracy, and this procedure iterates to sample the solutions with increasing quality. There are two ingredients that need to be specified in BO, i.e., the probabilistic prior over functions that expresses the assumptions of the function being optimized, and the acquisition function that determines the next solution to evaluate. For the probabilistic prior, BO usually uses the Gaussian process (GP)~\cite{book-gp4ml} prior because of its flexibility and analytical tractability. GP defines a distribution over functions specified by its mean function and covariance function. The function being optimized is assumed to be drawn from a GP prior, and this prior as well as the sampled data induce a posterior over functions. The acquisition function $a(\cdot)$, that is constructed from the model posterior, determines which solution should be evaluated next via a proxy optimization $\theta_{\text{next}}=\argmax_{\theta} a(\theta)$. The popular choices of acquisition function are probability of improvement~\cite{PI-Kushner}, expected improvement~\cite{EI-Mockus}, GP upper confidence bound~\cite{UCB-SrinivasKKS-icml10,UCB-FreitasSZ-icml12} and so on. 
As BO can exploit the prior information about the expected return and utilize this knowledge to select new policies to execute, more and more works focus on applying BO to RL, and early works on this direction is already reviewed in~\cite{BO-early-tutorial}.

Wilson et al.~\cite{WilsonFT-jmlr14} proposed a novel Gaussian process covariance function to measure the similarity between policies using the trajectory data generated from policy executions. Compared with the traditional covariance functions that relate the policy parameters, the proposed covariance function that relates the policy behavior is more reasonable and effective. Furthermore, they also introduce a novel Gaussian process mean function that exploits the learned transition and reward functions to approximate the landscape of the expected return. The developed Bayesian optimization approach tailored to reinforcement learning could recover from model inaccuracies when good transition and reward models cannot be learned. Empirical results on a set of classic control benchmarks verify its effectiveness and efficiency.

Calandra et al.~\cite{CalandraSPD-icra14} applied Bayesian optimization to design the gaits and the corresponding control policies of a bipedal robot. Three popular acquisition functions, as well as the effect of fixed versus automatic hyper-parameter selection, are analyzed therein. 
In~\cite{CalandraSPD16}, they additionally formalize the problem of automatic gait optimization, discuss some widely-used optimization methods, and extensively evaluate Bayesian optimization on both simulated tasks and real robots. The evaluation demonstrates that BO is very suitable for robotic applications since it could search a good set of gait parameters with only a few amount of experimental trials. By comparing the different variants of BO algorithms, they observe that the GP upper confidence bound acquisition function performs the best among them.

Marco et al.~\cite{MarcoBHS0ST-icra17} proposed a Bayesian optimization algorithm that can adaptively select among multiple information sources with different accuracies and evaluation costs, e.g., experiments on a real-world robot and simulators. This BO algorithm exploits the prior knowledge from simulations via maximizing the information gain from each experiment, and automatically integrates the cheap but inaccurate information from simulations with the accurate but expensive real-world experiments in a cost-effective way. The empirical results show that using the prior model information from simulators can reduce the amount of data required to find out the desirable control policies. Letham and Bakshy~\cite{corr-abs-1904-01049} augmented the on-line experiments with the off-line simulator observations to tune the live machine learning systems on the basis of the multi-task Bayesian optimization~\cite{SwerskySA-nips13}. The empirical study on the live machine learning systems indicates that, by directly utilizing a simple and biased off-line simulator together with a small number of on-line experiments, the proposed method can accurately predict the on-line outcomes and achieve the substantial gains. And the empirical result is consistent with the theoretical findings of the multi-task Gaussian process generalization.

The work termed as the Bayesian functional optimization (BFO)~\cite{VienZT-aaai18} extends the BO methods to the functional policy representations, and its motivation is similar to~\cite{VienDC-acml17}. BFO models the function space as a reproducing kernel Hilbert space (RKHS)~\cite{book-gp4ml}, and introduces an efficient update of the functional GP as well as a simple optimization of the acquisition functional. BFO bypasses the problem of manually selecting the features used in the function approximations to define the parameter space, and thus relaxes the performance reliance on the selected parameter space. The experimental result on the RL task whose policies are modeled in RKHS shows that BFO is able to compactly represent the complex solution functions.

Eriksson et al.~\cite{ErikssonPGTP-nips19} proposed the trust region Bayesian optimization (TuRBO) method in order to tackle the high-dimensional RL problems. They notice that the difficulty of high-dimensional optimization in RL comes from the plentiful local optima and the heterogeneity of the objective function (e.g., the sparse rewards in RL may lead to the objective function being nearly constant in a large region of the search space). This difficulty makes the task of learning a global surrogate model challenging. Thus, TuRBO maintains a collection of simultaneous local probabilistic models. Each local surrogate model shares the same benefits with Bayesian probabilistic modeling, and at the same time enables the heterogeneous modeling of the objective function. In TuRBO, a multi-armed bandit strategy is adopted to globally allocate the samples among the trust regions. The empirical results show that TuRBO outperforms the compared BO, EAs, and stochastic optimization on the RL benchmarks.

\subsection{Classification based model parameter updating}
The classification-based derivative-free optimization methods~\cite{EDAbook-06,yu.qian.racos,HashimotoYD-18,ZhouZSZ-aaai19} are recently proposed for non-convex functions with sample-complexity guarantees. The researchers notice that many derivative-free optimization methods are model-based, e.g., BO employs GP to model the function. The model-based optimization approaches learn a model from the evaluated solutions, and the model is then utilized to guide the sampling of solutions in the next iteration. The classification-based derivative-free optimization methods use a particular type of model, classification model from machine learning, to model the objective function. A classification model learns to classify the solutions in the search space into two categories, the good/positive and the bad/negative ones, according to the quality of the sampled solutions. The learned classifier partitions the search space into the good and bad regions. And then the solutions are sampled from the good regions with high probability. 

Yu et al.~\cite{yu.qian.racos} attempted to answer the crucial questions of classification-based derivative-free optimization, including which factors of a classifier effect the optimization performance, and which function class can be efficiently solved. They identify the critical factors, the error-target dependence and the shrinking rate, and propose the randomized coordinate shrinking (RACOS) classification algorithm to efficiently learn the classifier for both continuous and discrete search space. Given a set of good/positive and bad/negative solutions, RACOS learns an axis-aligned hyper-rectangle to cover all the positive but no negative solutions, and at the same time, the learning process is randomized and the hyper-rectangle is highly shrunk to meet the critical factors disclosed. However, RACOS needs to sample a batch of solutions in each iteration to update the classification model, while in RL, the environment often only offers the sequential policy evaluation. This means RACOS cannot be directly applied to policy search, where solutions are sampled sequentially. To address this issue, Hu et al.~\cite{hu2017sequential} further proposed a sequential version of RACOS called SRACOS, which is tailored to direct policy search in RL. SRACOS adapts the classification-based optimization for sequential sampled solutions by forming the sample batch via reusing the historical solutions. In each iteration, it only samples one solution, replaces a historical solution with the newly sampled one, and then updates the classifier as RACOS. They introduce three replacing strategies for maintaining the historical solutions, i.e., replacing the worst one, replacing by a randomized mechanism, and replacing the one that has the largest margin from the best-so-far solution. The empirical effectiveness is verified on the helicopter hovering task and controlling tasks in OpenAI Gym~\cite{MuJoCo-12,Atari2600-jair13,openai-gym}, and the neural network is used to represent the policy. The results indicate that SRACOS can surpass CMA-ES, CE, BO with respect to the total reward.

\section{Derivative-free model selection in reinforcement learning}
Deep reinforcement learning uses the deep neural networks to model the policy or the value function. The final determination of model depends on two aspects: the weights of connections in a neural network and the topology/architecture of a neural network. In the fourth section, we mainly review the cases when the topology of a neural network is fixed. Namely, the number of hidden layers, the number of hidden layer neurons, the edge set of connected neurons and the like are all fixed in advance. Under the condition of fixed neural network topology, the derivative-free methods are used to optimize the weights of connections to search for the optimal policy. 
However, relying on the fixed topology representation imposes some limitations, since it requires the user to correctly specify a good topology representation in advance. Unfortunately, in most tasks, the user cannot correctly guess the appropriate topology representation. Selecting an overly simple topology representation could result in the insufficient model expression ability. Even if the optimization algorithm can find out the optimal solution under this simple representation, the policy corresponding to the found optimal solution may still be unsatisfactory. On the other hand, selecting an overly complex topology representation could significantly improves the expressive power of the model, but the huge search space makes the optimization algorithm inefficient, and thus it is difficult to find out the optimal policy. Therefore, many works are devoted to developing the methods that can automatically finding out the appropriate topology representation of neural network. And the review of neuroevolution for learning a neural network architecture can be found from~\cite{yao1999evolving,stanley-neuroevolution-nature19}. 

This section mainly reviews the methods for optimizing neural network topologies using derivative-free optimization, where neural networks can represent the policy or the value function in reinforcement learning. The network topology and connection weights are variable at the same time, which further improves the expressiveness and flexibility of the reinforcement learning model. However, it challenges the derivative-free optimization algorithms. The search space becomes larger and more complex, and the evaluation of model quality becomes more time consuming and labor intensive.

The earliest and simplest work to optimize the neural network topologies using evolutionary algorithms might be traced back to the structured genetic algorithm (sGA)~\cite{Dasgupta1992}. sGA uses a two-part representation to describe each neural network. The first part represents the connectivity of the neural network in the form of a binary matrix. In this binary matrix, the rows and columns correspond to the neuron nodes in the network, and the value of each element in the matrix indicates whether there is an edge connecting a given pair of nodes. The second part represents the weight of each edge in the neural network. Via evolving these binary matrices with the connection weights, sGA can automatically discover the appropriate the network topology. However, sGA still has some restrictions. When a new topology is introduced (e.g., adding a new edge), the quality of the corresponding solution may be poor due to the fact that the weight associated with the new topology has not been optimized. Even if the topology ultimately corresponds to a better policy.

\begin{figure}[!bp]
	\centering\includegraphics[width=0.60\textwidth]{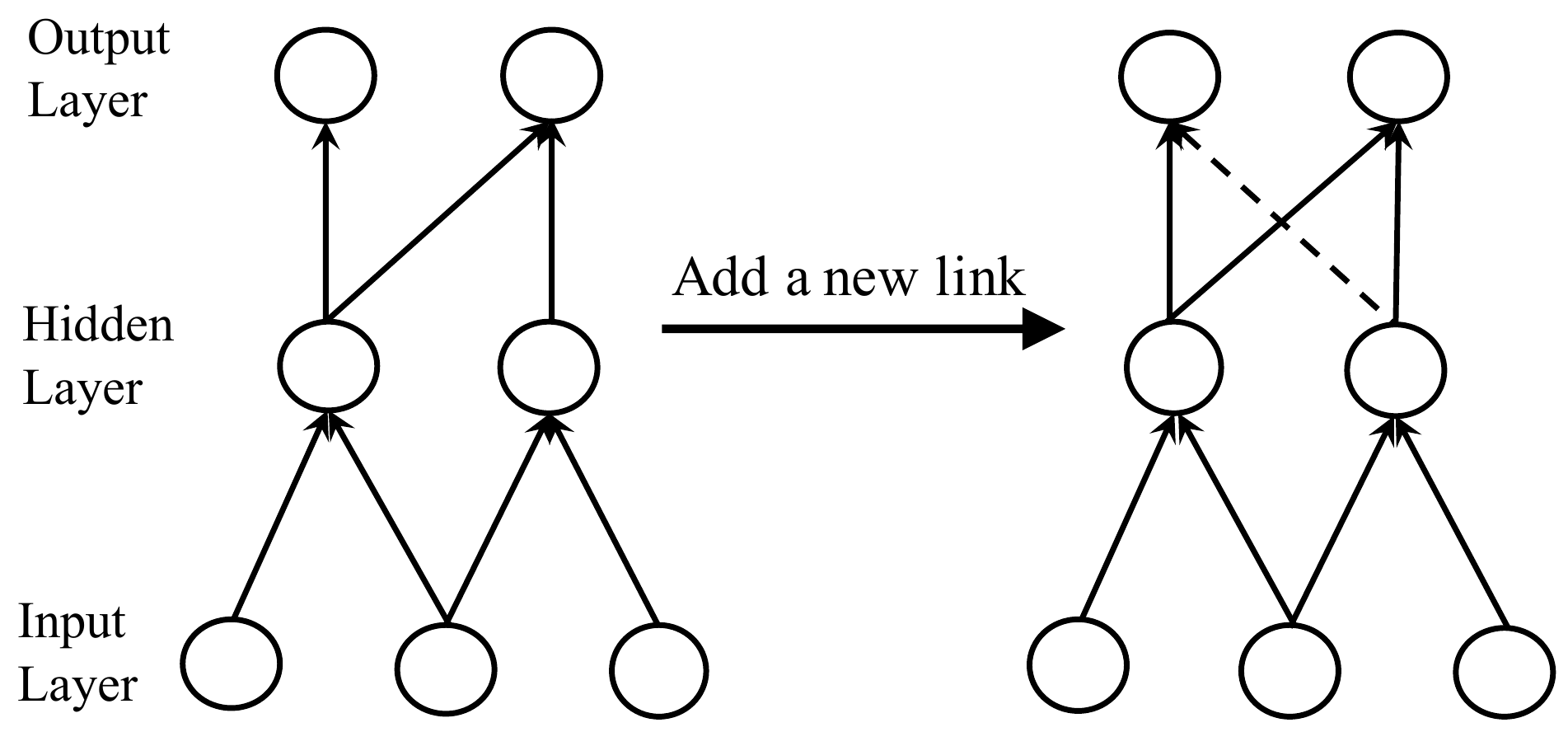}
	\caption{A new link (edge) is added between two existing nodes in NEAT~\cite{StanleyM-gecco02a,Stanley2002}.}\label{neat}
\end{figure}

Stanley and Miikkulainen~\cite{StanleyM-gecco02a,Stanley2002} proposed the neuroevolution of augmenting topologies (NEAT), which is a well-known approach of topology and weight evolving artificial neural networks (TWEANNs), and achieved a significant performance gains in RL. To represent networks of different topologies, NEAT uses a flexible genetic coding. Each network is described by an edge list, and each edge describes the connection between two neuron nodes and the weight. In NEAT, the architecture is evolved in a way of incremental growth from minimal structure. New architecture is added incrementally as structural mutations occur, and only those architectures survive that are found to be useful. During mutation, one can add new nodes or new connections to the network, as shown in Figure~\ref{neat}. To avoid the catastrophic crossovers, NEAT introduces the innovation numbers to record and track the historical origin of each individual. Whenever a new individual emerges through mutation, it receives a unique innovation number that belongs to itself. Therefore, the innovation numbers can be regarded as the chronology of all individuals produced during the evolutionary process. Experimental result on the pole-balancing benchmark shows that NEAT can be faster and better than the compared fixed-topology methods therein. 

Taylor et al.~\cite{TaylorWS06} conducted a detailed empirical comparison between NEAT and Sarsa~\cite{SinghS-ml96} in the Keepaway RL benchmark based on robot soccer. Sarsa is a kind of temporal difference~\cite{sutton_reinforcement_1998} methods that learn a value function to estimate the expected total reward for taking a particular action given a state. The results show that NEAT can learn better policies than Sarsa, but NEAT requires more fitness evaluations to achieve so. The results on two variations of Keepaway show that Sarsa can learn better policies when the task is fully observable, and NEAT can learn faster when the fitness function is deterministic. 
Whiteson and Stone~\cite{WhitesonS-aaai06,WhitesonS-jmlr06} enhanced the sample efficiency of evolutionary value function approximation via combining NEAT and a temporal difference method. The proposed algorithm can automatically discover the appropriate topologies for pre-trained neural network function approximators, and exploit the off-policy nature of a temporal difference method. 

Kohl and Miikkulainen~\cite{KohlM-nn09} noticed that NEAT could perform poorly on the fractured problems, where the correct action varies discontinuously as the agent moves from state to state. They introduce a method to measure the degree of fracture by utilizing the concept of function variation, and propose RBF-NEAT and Cascade-NEAT to improve the performance on the fractured problems through biasing or constraining the search for network topologies towards the local solutions. 
Gauci and Stanley~\cite{hyperNeat-aaai08} proposed the hypercube-based neuroevolution of augmenting topologies (HyperNEAT) that is able to exploit the geometric regularities in the two-dimensional game screen. Hausknecht et al.~\cite{hyperNeatGGP-gecco12} introduced a HyperNEAT-based general game playing (HyperNEAT-GGP) method to play Atari games. HyperNEAT-GGP reduces the learning complexity from the raw game screen by using a game-independent visual processing hierarchy aimed to identify the objects and entities on the screen. And the identified ones are input of HyperNEAT. The effectiveness of HyperNEAT-GGP is verified on Asterix and Freeway.

Ebrahimi et al.~\cite{Ebrahimi2017} introduced an approach to learn the control policy for the autonomous driving task aimed to minimize crashes and safety violations during training. The proposed approach learns to generate an optimal network topology from demonstrations by utilizing a new reward function that simultaneously optimizes model size and accuracy. They use a recurrent neural network to sequentially generate the description of layers of an architecture from a given design space, which is inspired by~\cite{ZophL-iclr17}. The variable length architectures are searched by an enhanced evolution strategy with a modification in noise generation. Finally, by combining this derivative-free policy search with demonstrations, the proposed approach can learn a policy that adapts to the new environment based on the rewards in the target domain. The experimental result shows that, when the agent learns to drive in a real simulation environment, this approach can learn more safely than the compared baseline method and has fewer cumulative crash times in the life cycle of the agent. 
Gaier and Ha~\cite{GaierH-nips19} attempted to answer the question of how important are the weight parameters of a neural network compared to its architecture. To deemphasize the importance of weights in the neural networks, they assign a single shared weight parameter to every network connection from a uniform random distribution, and only search for the architectures that perform well on a wide range of weights without explicit weight training. The proposed search method is inspired by NEAT~\cite{StanleyM-gecco02a,Stanley2002}. The empirical result shows that this method is able to find the minimal neural network architectures that perform well on a set of continuous control tasks only with a random weight parameter.

\section{Derivative-free exploration in reinforcement learning}
In most cases, RL algorithms share the exploration-learning framework~\cite{Yu18-ijcai}. An agent explores and interacts with an unknown environment to learn the optimal policy that maximizes the total reward from the exploration samples. The exploration samples involve states, state transitions, actions and rewards. Exploration is necessary in RL. Because achieving the best total reward on the current trajectory samples is not the ultimate goal, and the agent should visit the states that have not been visited before so as to collect better trajectory samples. This means that the agent should not follow the current policy tightly, and thus the exploration mechanisms need to encourage the agent to deviate from the previous trajectory paths properly and drive it to the regions with uncertainty. 
Derivative-free RL is naturally equipped with the exploration strategies. In the search process of derivative-free optimization methods, the designed mechanisms for sampling solutions and rules for updating model always consider the exploration. To name a few, the mutation and crossover operators in GAs help to explore the solution space, the GP upper confidence bound in BO uses variance to drive the search direction towards the regions with uncertainty, and the classification-based optimization algorithms usually adopt the global sampling to realize the exploration. 
Thus, derivative-free optimization methods can take part of the duty of exploration for RL when updating the policy or value function models from samples. Recently, some problem-dependent derivative-free exploration methods that could improve the sample efficiency have been proposed.

Conti et al.~\cite{conti2018improving} proposed the NSR-ES method, which attempted to enhance the ability of exploration in evolution strategies for deep RL. NSR-ES uses the novelty search (NS)~\cite{lehman_abandoning_2011} to explore. NS encourages policies to engage in different behaviors than those previously seen. The encouragement of different behaviors is realized through computing the novelty of the current policy with respect to the previously generated policies and then pushing the population distribution towards the regions in parameter space with high novelty. NS is hybridized with ES to improve its performance on sparse or deceptive deep reinforcement learning tasks. The proposed NSR-ES algorithm has been tested in the MuJoCo and Atari~\cite{MuJoCo-12,Atari2600-jair13,openai-gym}, and the overall performance is superior to the classic evolution strategies. 
Lehman et al.~\cite{safe-mutation-gecco18} noticed that the simple mutation operators, e.g., Gaussian mutation, may lead to the catastrophic consequences on the behavior of an agent in deep RL due to the drastic differences in the sensitivity of parameters. Therefore, they propose a set of safe mutation operators that facilitate exploration without dramatically varying the network behaviors. The safe mutation operators scale the degree of mutation of each individual weight according to the sensitivity of the outputs of network to that weight. And it is realized by computing the gradient of outputs with respect to the weights. 

Chen and Yu~\cite{ChenY-aamas19} considered the exploration as an extra optimization problem in reinforcement learning, and realized the exploration component guided by a state-of-the-art classification-based derivative-free optimization algorithm, rather than directly applying derivative-free optimization to both exploration and learning. The proposed method searches for the high-quality exploration policy globally by setting the performance of the sampled trajectories as the fitness value of that exploration policy. The target policy is optimized by the deep deterministic policy gradient (DDPG) algorithm~\cite{lillicrap_continuous_2015} on those explored trajectories. This method could overcome the sample inefficiency in derivative-free optimization and the exploration locality in gradient-based one. 
Vemula et al.~\cite{Vemula0B-aistats19} conducted a theoretical study on when exploring in the parameter space is better/worse than exploring in the action space by the black-box optimizer. They reveal that, when the dimensionality of action space and the horizon length are small, as well as the dimensionality of parameter space is large, exploration in the action space is preferred. Otherwise, when the horizon length is long and the dimensionality of policy parameter space is low, exploration in the parameter space is preferred.

\begin{figure}[!tp]
	\centering\includegraphics[width=0.45\textwidth]{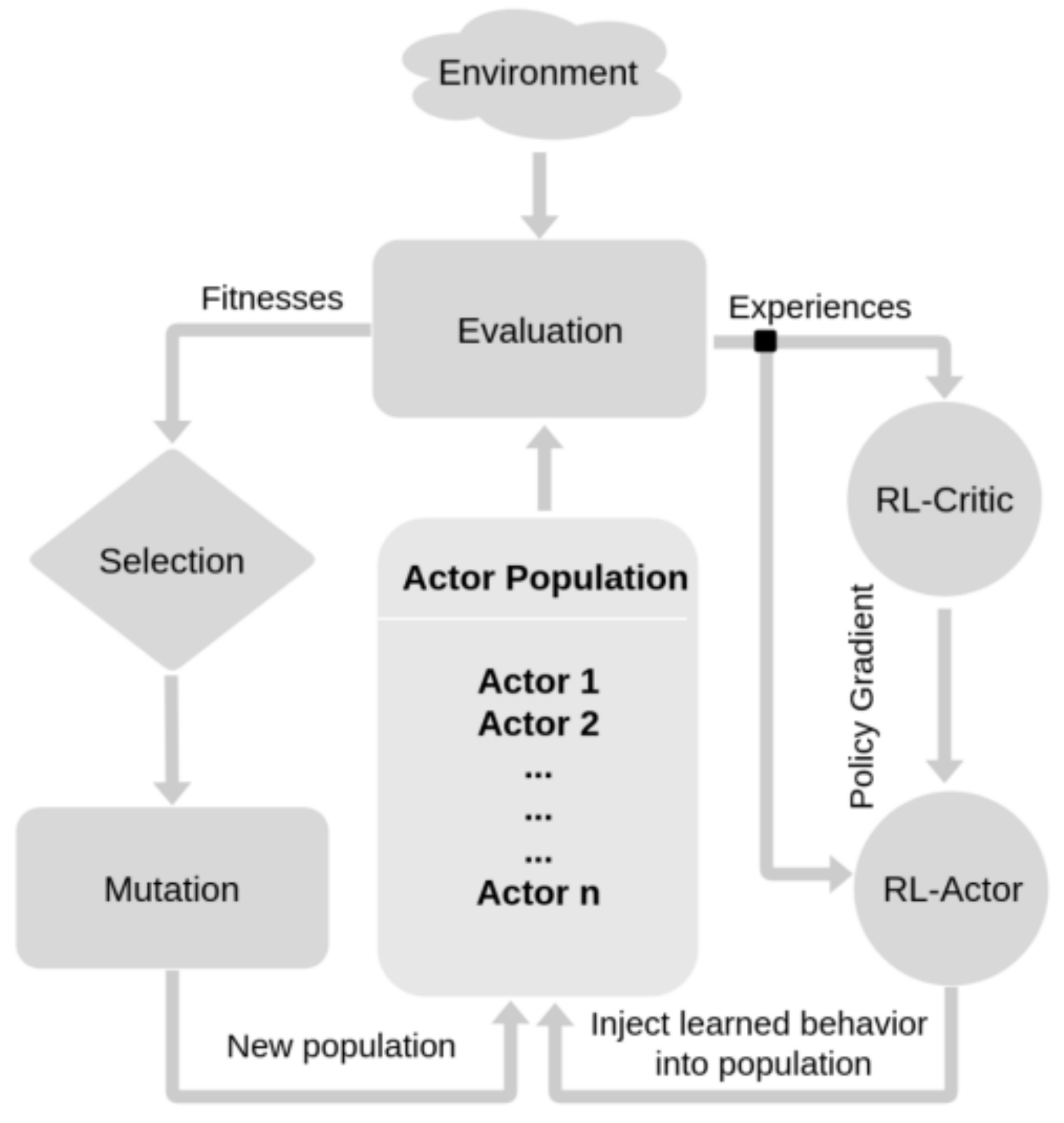}
	\caption{Figure from~\cite{khadka_evolution-guided_2018} that illustrates the structure of ERL.}\label{ERL}
\end{figure}

\begin{figure}[!tp]
	\centering\includegraphics[width=0.95\textwidth]{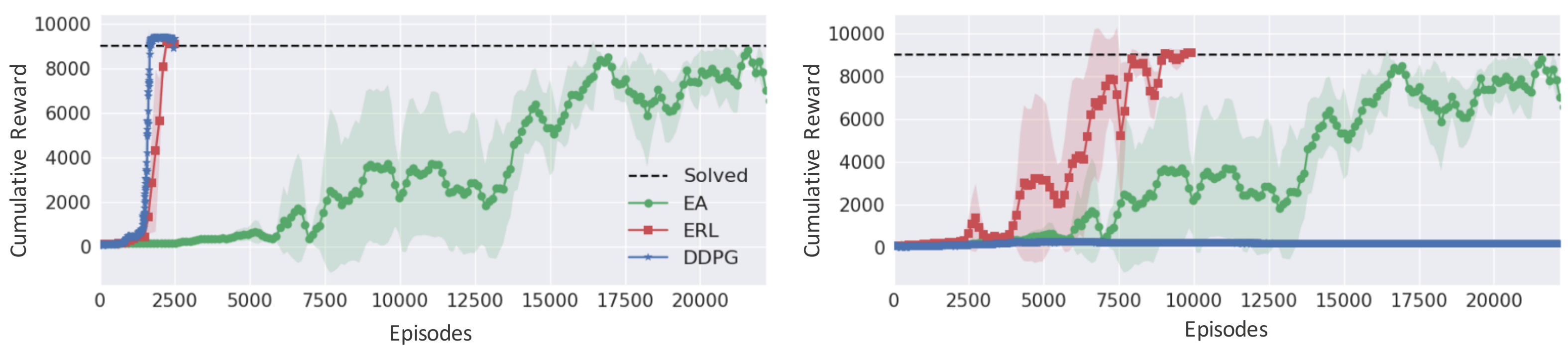}
	\caption{Figure from~\cite{khadka_evolution-guided_2018} that compares the performance (measured by the cumulative reward) of the proposed ERL, evolutionary algorithm, and DDPG on the standard (left sub-figure) and hard (right sub-figure) inverted double pendulum task.}\label{compare}
\end{figure}

Khadka and Tumer~\cite{khadka_evolution-guided_2018} proposed the evolutionary reinforcement learning (ERL) method. The core idea behind ERL is to combine the genetic algorithm (GA) and DDPG algorithm~\cite{lillicrap_continuous_2015}. The algorithmic flow is illustrated in Figure~\ref{ERL}. The algorithm can be divided into the genetic algorithm module and the reinforcement learning module. In the genetic algorithm module, ERL generates $n$ actors for parallel sampling, accumulates the cumulative reward of the sampled trajectories as the fitness of these $n$ actors, and then constructs $n$ new actors according to the mutation process of the genetic algorithm for the next sampling. 
In the reinforcement learning module, ERL collects samples of $n$ actors into the experience replay buffer, uses the policy of reinforcement learning algorithm itself to perform additional sampling, and adds the sampling results to the experience replay buffer. Then, according to the DDPG algorithm which is a gradient-based one, a mini-batch sample with a fixed number of rounds is selected from the experience replay buffer to optimize the policy.
At last, in order to inject the gradient information of policy into the genetic algorithm process, ERL periodically uses the policy to replace the policy with the worst fitness value among $n$ actors before the genetic algorithm performs further sampling.
ERL is tested on multiple environments in MuJoCo~\cite{MuJoCo-12,openai-gym}, and the result shows that ERL is superior to the compared derivative-free algorithms and gradient-based ones with respect to both algorithmic effectiveness and sample efficiency. For instance, as shown in Figure~\ref{compare}, we can observe that DDPG can easily solve the standard inverted double pendulum task, while fails on the hard one. Both tasks are similar for the evolutionary algorithm (EA). ERL is able to inherit the merits of both DDPG and EA, and successfully solves both standard and hard tasks similar to EA while utilizing the gradients for better sample efficiency similar to DDPG.

Colas et al.~\cite{colas_gep-pg_2018} proposed the GEP-PG algorithm that splits reinforcement learning into two parts: global exploration and policy gradient. In the global exploration process (GEP), it uses a method similar to novelty search to generate a variety of exploration policies based on the policy behavior space. Specifically, the GEP process samples a batch of policy behavior features from the behavior space. The behavior space is an artificially defined mapping from the policy trajectory to a vector, which is used to represent the type of a policy. It then uses the neighborhood method to find the clustering center for each policy behavior. After that, each clustering center is perturbed to generate a new set of exploration policies. After the new exploration policy is sampled, the behavior features of the exploration policy are marked, and according to the newly added marks, the cluster centers are re-generated by the neighborhood method. This process is repeated so that the generated exploration policies can be placed in different goals as much as possible, which could ensure the diversity of the generated policies. 
In the policy gradient (PG) process, it combines with the deep deterministic policy gradient (DDPG) algorithm~\cite{lillicrap_continuous_2015}, and adds the samples collected by the GEP process to the experience replay buffer. PG updates the policy by the fixed mini-batch. 
The GEP-PG approach enhances the diversity of exploration policies and is tested on the continuous mountain car and half-cheetah environments in MuJoCo~\cite{MuJoCo-12,openai-gym}. GEP-PG shows the better performance than the compared algorithms. 
Notably, some of the aforementioned methods, e.g., GEP-PG and ERL, are essentially a combination of derivative-free and gradient-based optimization. The derivative-free ones are used to better explore, and the gradient-based ones are used to better exploit.

\section{Parallel and distributed derivative-free reinforcement learning}
Although derivative-free methods could bring some good news to RL with respect to optimization and exploration, they mostly suffer from low convergence rate. Derivative-free optimization methods often require to sample a large amount of solutions before convergence, even if the objective function is convex or smooth~\cite{nips-JamiesonNR12,tit-DuchiJWW15,BachP16-colt16}. And the issue of slow convergence becomes more serious as the dimensionality of a search space increases~\cite{tit-DuchiJWW15,aaai-QianYu16}. Obviously, this issue will block the further application of derivative-free methods to RL. Fortunately, many derivative-free optimization methods are population-based. A population of solutions is maintained and improved iteratively. This characteristic makes them highly parallel. Thus, derivative-free optimization methods can be accelerated by parallel or distributed computation, which alleviates their slow convergence. Furthermore, for parallel/distributed derivative-free methods, the data communication cost is lower compared with gradient-based ones, since only one-dimensional scalars (fitness values) instead of gradient vectors or Hessian matrices need to be conveyed. This merit could further compensates for the low convergence rate. 

The ES method~\cite{salimans2017evolution} mentioned in the fourth section is a typical parallel case of the derivative-free method. Moreover, by the virtue of high parallelization property of ES, a novel communication strategy based on common random numbers is introduced to further reduce the communication cost, and it can make the algorithm well scaled to a large number of parallel workers. It takes only 10 minutes to reach 6000 long-term reward in the 3D Humanoid environment with 1440 cores. And with the increasing of core amount, the consumption of training time shows a relatively stable linear decline. 
SRACOS~\cite{hu2017sequential} is a classification-based derivative-free optimization algorithm for direct policy search. Its distributed version ZOOpt~\cite{liu_zoopt_2017}, an open-source toolkit, is implemented by the Julia language. And the experimental result shows that the algorithm can support more than 100 processes for parallel computing.
The parallel strategies of ES and SRACOS are similar. Both are based on a certain generation module to construct multiple policies simultaneously. ES is based on Gaussian perturbation for current parameters, and SRACOS is based on a randomized coordinate shrinking classifier. Policies are run and their fitness values are evaluated in parallel. The generation module is updated by these fitness values, and the next batch of policies is constructed.

Reinforcement learning algorithms sometimes are sensitive to the hyper-parameters, such as learning rate and entropy penalty coefficient. Finely tuned hyper-parameters can accelerate learning and improve the performance of policy. Previous hyper-parameter adjustment methods are parallel search algorithms, which use the performance of final policy under different hyper-parameters to adjust hyper-parameters. Such methods require a large amount of computational resources to simultaneously optimize multiple models. 
Population-based training (PBT)~\cite{jaderberg_population_2017} is a framework to simultaneously optimize model parameters and adjust hyper-parameters with a derivative-free method. The method is shown in Algorithm~\ref{framework}. The $\text{step}(\theta\,|\,h)$ is a function to conduct model optimization according to the current hyper-parameters $h$ and the model parameters $\theta$. The optimization method depends on tasks (e.g., SGD for supervised learning and DDPG for reinforcement learning). The $\text{eval}(h)$ is a function to realize performance evaluation. The evaluation method also depends on tasks (e.g., the loss for supervised learning and the total reward of a policy for reinforcement learning). If the model evaluation result $p$ is better than a threshold, population-based evolution will start to adjust the hyper-parameters $h$. 
\begin{algorithm}[!tbp]
	%\begin{algorithm}[!b]%[!t]
	\caption{Population Based Training (PBT)~\cite{jaderberg_population_2017}}
	\label{framework}
	\begin{algorithmic}[1]
		\ENSURE \textsc{Train} ($\mathcal{P}$)\\
		\STATE initial population $\mathcal{P}$
		\FOR{$(\theta, h, p, t) \in \mathcal{P}$ (asynchronously in parallel)}
		\WHILE{not end of training}
		\STATE one step of optimization: $\theta \leftarrow \text{step}(\theta\,|\,h)$
		\STATE current model evaluation: $p \leftarrow \text{eval}(h)$
		
		\IF{$\text{ready}(p,t,\mathcal{P})$}
		\STATE use the rest of population to find better solution: $h', \theta' \leftarrow \text{exploit}(h, \theta, p, \mathcal{P})$
		\IF{$\theta\neq \theta'$}
		\STATE produce new $h$: $h, \theta \leftarrow \text{explore}(h', \theta', \mathcal{P})$
		\STATE new model evaluation: $p\leftarrow \text{eval}(\theta)$ 
		\ENDIF
		\ENDIF
		\STATE update $\mathcal{P}$ with new $(\theta, h, p, t + 1)$
		\ENDWHILE
		\ENDFOR
		\RETURN  $\theta$ with the highest $p$ in $\mathcal{P}$
	\end{algorithmic}
\end{algorithm}
There are two main functions in the population-based evolution, i.e., $\text{exploit}(h, \theta, p, \mathcal{P})$ and $\text{explore}(h', \theta', \mathcal{P})$. The $\text{exploit}(h, \theta, p, \mathcal{P})$ function exploits the best hyper-parameters $h$ and its model parameters $\theta$ to conduct evolution. For example, replacing the worst solution in $\mathcal{P}$ with the best one. The $\text{explore}(h', \theta', \mathcal{P})$ function explores the new unknown hyper-parameters $h$ based on the current population $\mathcal{P}$. For example, perturbing $h'$ with Gaussian noise to generate the new hyper-parameters $h$. After population evolution, the new hyper-parameters $h$ as well as its model parameters $\theta$ will be added to $\mathcal{P}$. 
The main characteristic of PBT is that the process of parameter optimization and hyper-parameters adjustment is hybrid, which can not only reduce the computational cost but also help algorithm adjust hyper-parameters in dynamic environments. PBT has succeeded in RL when combining with A3C~\cite{mnih_asynchronous_2016}. The empirical results in the DeepMind Lab 3D environment~\cite{BeattieLTWWKLGV16}, Atari games~\cite{Atari2600-jair13,openai-gym}, and the StarCraft~II environment tasks~\cite{abs-1708-04782} show that PBT increases the final performance of the agents when trained with the same number of episodes.

Ray~\cite{moritz_ray_2018}, a distributed framework proposed by the researchers from UC Berkeley, integrates the distributed implementation of the PBT algorithm. 
Online meta-learning by parallel algorithm competition (OMPAC)~\cite{elfwing_online_2018} is also a derivative-free distributed hybrid optimization algorithm in RL. This algorithm can be regarded as an improvement based on PBT. The main difference between OMPAC and PBT is the evolution mechanism. In OMPAC, all solutions are evaluated in a synchronized manner after a fixed number of samples, and OMPAC uses the stochastic universal sampling~\cite{baker_reducing_1987} to select solutions for continual learning. In PBT, solutions are evaluated asynchronously (e.g., after a fixed number of training steps). 
Jaderberg et al.~\cite{PBTapp-science19} demonstrated that the agent using only pixels and game points as input could learn to play highly competitively in a popular multi-player first-person video game. The 
ingredients in the proposed method include PBT of agents, internal reward optimization, and hierarchical RL with scalable computational architectures. 
Jung ea al.~\cite{JungPS-iclr20} proposed the population-guided parallel policy search (P3S) to improve the performance of RL. P3S employs a population to search a better policy by exploiting the best policy information which is similar to PBT. However, the way of using the best policy information is different. In P3S, instead of copying the parameter of the best learner, it introduces a soft manner to guide the population for better search in the policy space. P3S maintains a batch of identical learners with their own value-functions and policies sharing a common experience replay buffer, and searches a good policy cooperated by the guidance of the best policy information. In a nutshell, parallel and distributed computation is necessary for derivative-free reinforcement learning.

\section{Discussion and Conclusion}
In this article, we summarize some recent progress in applying derivative-free optimization to reinforcement learning. Since derivative-free optimization is a generally applicable tool, it can be utilized in different levels and aspects of reinforcement learning. Here, we focus on the aspects of parameter updating, model selection, exploration, and parallel/distributed derivative-free methods. Due to the complexity of policy search in reinforcement learning, successful research studies of using derivative-free optimization are noticeable in all of these aspects. 

Derivative-free reinforcement learning approaches have their own advantages. Firstly, during the optimization process, they do not perform gradient back-propagation, are easy to train, possess the ability of search globally, do not care whether the reward function is sparse or dense, do not care the length of time horizons, do not require to carefully tune the discount factor, and can be applied to optimize the non-differentiable policy functions. Secondly, they could provide better exploration tailored to the problems. Thirdly, they are highly suitable for parallelization and only require the low communication cost. Usually, the amount of information that needs to be exchanged among workers does not rely on the size of neural networks.

At the same time, we also notice that the combination of derivative-free optimization and reinforcement learning has several limitations. These are mainly from the issues of current derivative-free optimization methods. The foremost ones could be the issues of sample complexity and scalability.

Derivative-free optimization algorithms could be comparable to some state-of-the-art gradient-based ones in reinforcement learning tasks, but usually they require more samples~\cite{salimans2017evolution}. The low sample efficiency may block the further application of derivative-free optimization algorithms to reinforcement learning. Recently, some emerging works focusing on improving sample efficiency by the way of introducing sample reuse~\cite{abs-1808-05832,LiuZYBQYL-aaai19}, surrogate models~\cite{StorkZBE-gecco19}, importance sampling~\cite{BibiBSGR-aaai20}, and momentum~\cite{Chen0XLLHC-nips19,GorbunovBSBR-iclr20}. However, these advances are far from enough and more future works on this direction are urgently needed.

Due to the sampling mechanism, derivative-free optimization methods have limitation of scaling up to the search space with very high dimensionality. This is a crucial issue for deep reinforcement learning. Deep neural network models often have more than a million parameters, for which parameter optimization directly by derivative-free optimization can be inefficient. While some recent works dedicated to tackling the scalability issue~\cite{Kandasamy2015,wangziyu2016,aaai-QianYu16,Qian2016SRE,tec-YangTY18,nips-MutnyK18,MullerG-ppsn18,ErikssonPGTP-nips19,LiZLZ-tcyb20}, more future works on developing the scalable derivative-free optimization methods tailored to reinforcement learning are appealing.

Other issues include noise-sensitivity as well as structure-insensitivity. Since derivative-free optimization relies on solution evaluations, noisy evaluation can badly affect the policy search in reinforcement learning~\cite{Wang2017VS}. Reinforcement learning usually has inner structures that can help better solve the learning, such as hierarchical policy models and curiosity-driven exploration, while derivative-free methods commonly ignore the inner structure of the problem. How to make the derivative-free optimization well aware of the inner structure and utilizing the structure is also an important direction and remains under explored. 

For the above issues, the hybrid of derivative-free and gradient-based algorithms could be a potential and promising way. Some existing successful attempts reviewed in this article indicate that, by fusing derivative-free and gradient-based algorithms in a clever manner, the strengths of both sides could be absorbed, and more powerful reinforcement learning algorithms would be expected.

With the successful cases and fast progress, we expect in a near future that derivative-free methods will play an even more important role in developing novel and efficient reinforcement learning approaches, and hope that this review article will serve as a catalyst for this goal.

\section*{Acknowledgments}
This article has been accepted by \href{http://journal.hep.com.cn/fcs/EN/2095-2228/current.shtml}{Frontiers of Computer Science} with DOI: 10.1007/s11704-020-0241-4 in 2020. This work is supported by the Program A for Outstanding Ph.D. Candidate of Nanjing University, National Science Foundation of China (61876077), Jiangsu Science Foundation (BK20170013), and Collaborative Innovation Center of Novel Software Technology and Industrialization. The authors would like to thank Xiong-Hui Chen and Zhao-Hua Li for improving the article. The authors also would like to thank the anonymous reviewers for their detailed comments that have led to a significant enhancement of this article.

\bibliographystyle{named}
\bibliography{FCS_survey_DFRL}

\end{document}